\documentclass[10pt]{article} 
\usepackage[preprint]{tmlr}

\usepackage{amsmath,amsfonts,bm}

\def\eqref#1{equation~\ref{#1}}

\def\1{\bm{1}}

\DeclareMathAlphabet{\mathsfit}{\encodingdefault}{\sfdefault}{m}{sl}
\SetMathAlphabet{\mathsfit}{bold}{\encodingdefault}{\sfdefault}{bx}{n}

\usepackage{hyperref}
\usepackage{url}
\usepackage{graphicx}
\usepackage{tabularray}
\usepackage{multirow}

\title{BRAT: Bonus oRthogonAl Token for Architecture Agnostic Textual Inversion}

\author{\name James Baker \email jlbaker361@gmail.com \\
      \addr Department of Computer Science\\
      University of Maryland, Baltimore County}

\def\month{07}  
\def\year{2024} 
\def\openreview{\url{https://openreview.net/forum?id=XXXX}} 

\begin{document}

\maketitle

\begin{abstract}
Textual Inversion remains a popular method for personalizing diffusion models, in order to teach models new subjects and styles. We note that textual inversion has been under-explored using alternatives to the UNet, and experiment with textual inversion with a vision transformer. We also seek to optimize textual inversion using a strategy that does not require explicit use of the UNet and its idiosyncratic layers, so we add bonus tokens and enforce orthogonality. We find the use of the bonus token improves adherence to the source images and the use of the vision transformer improves adherence to the prompt. Code is available at \url{https://github.com/jamesBaker361/tex_inv_plus}.
\end{abstract}

\section{Introduction}
When the British pop singer Charli XCX sings \textit{"When you're in the mirror, do you like what you see? When you're in the mirror, you're just looking at me"} in her song 360 \citep{charli2024}, she is implicitly acknowledging that mirrors show an instance of a subject. However, due to her eminence as an icon, the specific instance is replaced with her. Charli XCX's lyrics imply that it is very significant and meaningful \textit{which} person appears in the mirror when the audience looks in the mirror. Depicting a generic picture of a cat is less useful than depicting a \textit{specific} cat. The pop singer herself has admitted that she doesn't just wear any pair of boots but strongly prefers a very specific model of black Prada boots \citep{gq2024}. This affinity towards specific instances of subjects motivates the personalization of text-to-image models. 

While most of the standard pre-trained diffusion models know a few specific instances of specific characters or objects (for example, most off-the-shelf diffusion models can generate images of the Statue of Liberty, not just a generic statue when prompted to do so), there has been no shortage of clever  methods to expand their output space to a new subject or style \citep{ruiz2022dreambooth, li2023blipdiffusion,ma2023subjectdiffusionopen,purushwalkam2024bootpig,wang2024instantid,ye2023ipadapter, chen2024idalignerenhancingidentitypreservingtexttoimage,zhang2023realworld}  We revisit one of the earliest methods of personalizing diffusion models, known as Textual Inversion \citep{gal2022image}. Textual inversion is most often used to teach diffusion models new subjects. However,  it can also be used effectively to teach new styles, similar to style transfer, but without a content image. We also note that most, if not all, textual inversion literature is wedded to the use of the UNet, and many of the optimizations are UNet-specific. While not all diffusion models use UNet the massive size of the text encoders used with other models such as Vision Transformers \citep{dosovitskiy2021image} can make textual inversion on them difficult. However, this problem can be circumvented with the use of adapters \citep{zhao2024bridging} that map the embeddings of one text encoder into the space of another, opening up the possibility of textual inversion on Vision Transformers. At the same time, we see that the fixation on UNet architecture has meant a dearth of model-agnostic improvements to textual inversion. Given the state of the field, our contributions are as follows:
\begin{itemize}
    \item We apply textual inversion to a non-UNet architecture
    \item We use a new token method, which we call BRAT, that is agnostic to choice of denoising model
    \item We demonstrate that BRAT improves adherence to the source image, and the non-UNet architecture improves human preference rating and prompt adherence
\end{itemize}

\section{Related Work}
\subsection{Text to Image}
The Generative Adversarial Network \citep{goodfellow2014generative} was a milestone in using machine learning to create new visual artifacts. Later works like Deepdream \citep{mordvintsev2015inceptionism} and style transfer using Gram matrices \citep{gatys2015neural} or CycleGAN \citep{zhu2020unpaired} could generate images conditioned on source images. GANs were also implemented conditioned on text \citep{mirza2014conditional,reed2016generative}, allowing users even more control of what these models could create. Recently, autoregressive methods \citep{ramesh2021zeroshot,ding2021cogview,yu2022scaling} and  diffusion \citep{sohldickstein2015deep, ho2020denoising,podell2023sdxl} have emerged and produce even higher quality images than prior works. The latter method usually consists of a denoising model, most commonly the Unet \citep{ronneberger2015unet}, conditioned on text embeddings from a pretrained text encoder, progressively tries to guess the amount of random noise in a corrupted image or latent embedding of one, however the "denoising" task can predict other forms of image corruption as well \citep{bansal2022colddiffusioninvertingarbitrary,liu2024museummakercontinualstylecustomization}.



\subsection{Textual Inversion}
Foundation models are trained on massive datasets. For example, stable diffusion \citep{rombach2022highresolution} was trained on a few billion images \citep{schuhmann2022laion5b}; DALLE-3 \citep{ramesh2023dalle3} was supposedly trained on the data used for CLIP \citep{ramesh2021zeroshot} and DALL-E \citep{radford2021learning}, each containing hundreds of millions of images. Subsequently, these models can accurately portray an extremely wide range of concepts. However, they do not have the ability to consistently generate instances of a specific subject that may be similar to things but not necessarily found in the training data. Emerging almost simultaneously, Dreambooth \citep{ruiz2022dreambooth} tuned the Unet, while Textual Inversion \citep{gal2022image} tuned a new token in the text encoder vocabulary to teach these models unique new concepts and optimized the embedding to match a few example images. There were many modifications and expansions to Textual Inversion. For example, DreamArtist \citep{dong2023dreamartist} included a "negative" token,  ProSpect \citep{zhang2023prospect} used different tokens for different timesteps of the diffusion process, P+ \citep{voynov2023p} and MATTE \citep{agarwal2023image} used different tokens for different layers of the UNet, and NeuralSpace \citep{alaluf2023neural} trained an auxiliary network to modify the token, conditioned on the UNet layer and timestep. Similar to textual inversion, some works focus on optimizing a prompt description of an image without necessarily introducing a new token into the vocabulary \citep{wen2023hardpromptseasygradientbased, yu2024uncovering}.

\section{Method}
\subsection{Baseline}
We describe standard textual inversion. Given:
\begin{itemize}
    \item Image dimensions Height \(H \in \mathbb{N}\), Width \(W \in \mathbb{N}\) and Channels \(C \in \mathbb{N}\)
    \item Image \(x \in \mathbb{R}^{H \times W \times C}\) drawn from real dataset \(\mathcal{D}\)
    \item Latent Height \(L_H <<< H \in \mathbb{N}\), Latent Width \(L_W <<< W \in \mathbb{N}\) and Latent Channels \(L_C \in \mathbb{N}\)
    \item Variational Autoencoder \citep{kingma2014auto} \(\mathcal{E} : \mathbb{R}^{H \times W \times C} \rightarrow \mathbb{R}^{L_H \times L_W \times L_C}\)
    \item Timestep \(t \in [0,T]\), where \(T\) is the maximum number of timesteps
    \item Time-conditioned noise \( \epsilon \sim \mathcal{N}(0,1 | t) \)
    \item Noised Latent Embedding \(z_t\), where \(z_t= \mathcal{E} (x) + \epsilon\)
    \item Text token embedding dimension \(M \in \mathbb{N}\), Text vocab size \(V \in \mathbb{N} \), \(M <<< V\), prompt length \(N \in \mathbb{N}\)
    \item Prompt \(p \in \textbf{text}\)
    \item Text Encoder \(c_{\phi} : \textbf{text} \rightarrow \mathbb{R}^{N \times M}\), parameterized by \(\phi\)
    \item Denoising model, \(\epsilon_{\theta}\) parameterized by \(\theta\)
    \end{itemize}
    The traditional Latent Diffusion Model denoising training objective is
    \[\mathcal{L}_{LDM}=\mathbb{E}_{z \sim \mathcal{E}(x), x \sim \mathcal{D}, t, p } || \epsilon - \epsilon_{\theta}(z_t, t, c_{\phi}(p))||_2^2\]
    Usually the \(\theta\) parameters are optimized. For textual inversion, \(\theta\) and \(\phi\) are frozen, and the only trainable parameter is the embedding of the new token. The new token is usually a pseudoword added to the vocabulary such as \textbf{"<sks>"}, and the embedding is represented as \(v^*\). The dataset is much smaller than the original dataset used to train the diffusion model, and is instead 3-5 pictures of the subject. The prompts are usually very basic, along the lines of \textbf{"a picture of <sks>"} or \textbf{"a photo of <sks>"}.
    
\subsection{Adapting Text Encoders to Different  Denoisers}
Traditionally, textual inversion finetunes off of a transformer and a UNet. Many of the optimizations have been based on leveraging unique features of the UNet, such as how different layers correspond to different attributes of an image \citep{voynov2023p,agarwal2023image,alaluf2023neural}. A challenge to using textual inversion with different combinations of text encoder and denoiser is larger text encoders can be harder to train; refer to appendix \ref{appendix-fail} for some examples. Work by \citet{zhao2024bridging} allows us to use a pretrained adapter module that maps the embeddings of one text encoder into the space of another. For example, the PixArt-\(\alpha\) vision transformer \citep{chen2023pixartalpha} architecture has impressive visual results, but uses a text encoder of a few billion parameters. Training and optimizing this text encoder with textual inversion requires more memory, and thus more expensive hardware. However, using an adapter, we can perform textual inversion on a smaller model of only a few million parameters.
\subsection{BRAT: Bonus Orthogonal  Token}
Given that many of the textual inversion improvements are UNet-specific, we would like to try a token strategy that is agnostic to our choice of denoiser. We also investigate if one token may be insufficient to capture all the relevant information of one concept. Many concepts are better described with multiple words than just one; "orange cat" is more descriptive than "cat". We add an auxiliary pseudoword \textbf{"<fkf>"}, which we refer to as the "bonus" token, and corresponding embedding \(w^{*}\). However, we don't want \(w^*=v^*\) or \(w^*=-v^*\), as that essentially means \(w^*\) and \(v^*\) embed the exact same information. We want to encourage the two embeddings to be orthogonal, so we introduce a new regularization term:
\[\mathcal{L}_{Spare}= \lambda  [\textbf{cos}(w^*,v^*)^2]\]
Where \textbf{cos} = cosine similarity and \(\lambda\) is a scalar weight coefficient, which we choose to be 0.01. This loss penalizes cosine similarity of 1 and -1 the same, thus discouraging \(w^*=v^*\) and \(w^*=-v^*\) and penalizes a cosine similarity of 0, encouraging \(w^*\) and \(v^*\), to be orthogonal and capture different aspects of the subject. The prompts follow the format of baseline textual inversion,  such as \textbf{"a picture of <sks> <fkf> "} or \textbf{"a photo of <sks> <fkf>"}. The objective becomes
\[\mathcal{L}=\mathcal{L}_{Spare} + \mathcal{L}_{LDM}\]
Where the only tunable parameters are the embeddings \(w^*,v^*\). We can then expand this to any number of bonus tokens, all of which are trained to be orthogonal to each other and the initial pseudoword. For our experiments we try with one and three bonus tokens. We call our method \textbf{BRAT}, for \textbf{B}onus o\textbf{R}thogon\textbf{A}l \textbf{T}oken.
\section{Experiments}
\subsection{Datasets}
Following \citet{gal2022image}, we used textual inversion to teach new subjects and styles. All images were converted to RGB, padded to be square, resized to \(512 \times 512 \times 3\) using bilinear interpolation and normalized to be between \([-1,1]\).
\subsubsection{Subject Data}
For the subject data, \(D_{sub}\), we used the 30 non-human subjects (cans, dogs, toys) from the original Dreambooth paper \citep{ruiz2022dreambooth}, retrieved from \url{https://github.com/google/dreambooth}. We only used three images for each subject, even though many of the images in the source repository had more than three images. Each subject was already labeled. We used the embedding of the subject label to initialize the custom placeholder token and spare token embeddings. Sometimes, this subject label had multiple words or extraneous numbers (for example, colorful\_sneaker or cat2). In that case, we removed any numbers and/or only used the embedding of the second word (fancy\_boot became boot, dog3 became dog, etc.). 
\subsubsection{Style Data}
For the style data, \(D_{sty}\), we selected  sixteen artists off of \url{deviantart.com}, and chose an image from each one that we felt reflected their unique artistic style. Refer to appendix \ref{appendix-style} to see them. New token embeddings were simply initialized with the embedding of the word "art". We should caveat that the task for the style dataset is \textit{not} style transfer; there is no content image that we are attempting to imbue a style unto; we want to generate a \textit{new} image with the stylistic features of the source image.

\subsection{Metrics}
We used six different metrics for evaluation:

\begin{itemize}
     \item \textbf{CLIP Similarity (CLIP Sim):} CLIP \citep{radford2021learning} Image encodings do not disentangle content and style but are the standard way of condensing images into a vector space. For both datasets, for each source image, we found the average cosine distance between the \textit{CLIP} embedding of each source image and each validation image and reported the average distance across subjects.
    \item \textbf{CLIP Consistency (CLIP Cons):} For both datasets, given the CLIP embeddings of each validation image, we calculated the distance between each pair and averaged them to get the consistency score
    \item \textbf{Style/Content Similarity (Style/Cont Sim):} Previous works \citep{tumanyan2022splicing,kwon2023diffusionbased} have found that using the activations of the intermediate layers of the vision transformer loaded from the \textbf{dino-vits16} checkpoint \citep{caron2021emerging} are good embeddings of the content of an image. For each subject, we found the average cosine distance between the \textit{content} embedding of each source image and each validation image and reported the average distance across subjects.
    For each style, we found the average cosine distance between the \textit{style} embedding of the single source image and each validation image and reported the average distance across subjects.
    \item \textbf{Style/Content Consistency (Style/Cont Cons):} For each subject, given the content embeddings of each validation image, we calculated the distance between each pair and averaged them to get the consistency score. For each style, given the style embeddings of each validation image, we calculated the distance between each pair and averaged them to get the consistency score. This metric measured how consistent the representation of the subject was across different prompts.
   
    \item \textbf{Image Reward (Img Rew):} In order to approximate subjective human preferences on whether an image is "good" or not, we used the pretrained Image Reward model \citep{xu2023imagereward} downloaded from the python package \textbf{image-reward} and used the \textbf{ImageReward-v1.0} checkpoint to score each validation image, and averaged them for this metric.
    \item \textbf{Prompt Similarity (Pro Sim):} The CLIP model is multimodal and can embed texts and images into the same space. So, for each validation prompt and subsequent generated image, we can extract embeddings of both prompt and image and find the cosine similarity to compute how close to the prompt the image is. We found the cosine similarity between each validation image and its source prompt and averaged them for this metric.
    
\end{itemize}
\subsection{Models}
We used 2 types of denoising model: a UNet, specifically the \textbf{stable-diffusion-v1-4} checkpoint from \url{https://huggingface.co/CompVis/stable-diffusion-v1-4}, and PixArt-\(\alpha\) vision transformer \citep{chen2023pixartalpha}, based off of \citet{peebles2023scalable}, specifically the \textbf{PixArt-XL-2-512x512} checkpoint from \url{https://huggingface.co/PixArt-alpha/PixArt-XL-2-512x512}. By default, the UNet uses a CLIP encoder, specifically the transformer-based text encoder used in \textbf{clip-vit-large-patch14}, downloaded from \url{https://huggingface.co/CompVis/stable-diffusion-v1-4} (but identical to the checkpoint from \url{https://huggingface.co/openai/clip-vit-large-patch14}). The vision transformer uses an extremely large T5 encoder, specifically the \textbf{4.3B Flan-T5-XXL} checkpoint downloaded from \url{https://huggingface.co/PixArt-alpha/PixArt-XL-2-512x512}, respectively. We found that training the vision transformer with the \textbf{4.3B Flan-T5-XXL} encoder was extremely slow to converge even with a higher learning rate and more training epochs, so we did not explore this. Refer to appendix section \ref{appendix-fail} for some visual examples.

However, we also leverage the pretrained adapters \citep{zhao2024bridging} from \url{https://huggingface.co/shihaozhao/LaVi-Bridge} so that we can use alternative text encoders. This repository contained only three adapters, adapting a \textbf{t5-large} encoder to a unet, adapting a \textbf{t5-large} encoder to a vision transformer and adapting a llama encoder to a unet.  We used the \textbf{t5-large} checkpoint from \url{https://huggingface.co/google-t5/t5-large} (which is a few million parameters, unlike the other T5 encoder) for both the UNet and vision transformer. We experimented with the \textbf{Llama-2-7b-hf} checkpoint \citep{touvron2023llama}, but we found this often failed to learn the target concept, just like the larger transformer encoder. Refer to appendix section \ref{appendix-fail} for some visual examples. This gives us three different combinations of noise predictor and text encoder, as detailed in table \ref{tab:models}. 
\begin{table}[h]
    \centering
    \begin{tabular}{|c|c | c |c|}
    \hline
         Name & Text Encoder & Denoiser & Uses Adapter? \\
         \hline
         T5 Trans & t5-large & PixArt-XL-2-512x512 & \(\checkmark\) \\
         T5 UNet & t5-large & stable-diffusion-v1-4 & \(\checkmark\) \\
         CLIP UNet & clip-vit-large-patch14 & stable-diffusion-v1-4 & \(\times\) \\
         \hline
    \end{tabular}
    \caption{Caption}
    \label{tab:models}
\end{table}

We tested a few token strategies for each model:
\begin{enumerate}
    \item \textbf{Default:} identical to \citet{gal2022image}.
    \item \textbf{Multi 10:} based off of ProSpect \citep{zhang2023prospect}, we have ten tokens, each corresponding to five inference steps 
    \item \textbf{Multi 50:} based off of ProSpect \citep{zhang2023prospect}, we have a separate token for each inference step, for a total of 50 tokens.
    \item \textbf{Negative:} based off DreamArtist \citep{dong2023dreamartist}, we use a negative token \(p_{-}\), and the loss is \(||\epsilon - f(\epsilon_{\theta}(z_t, t, c_{\phi}(p)), \epsilon_{\theta}(z_t, t, c_{\phi}(p_{-})))\), where \(f(a,b)=b+ \gamma (a-b)\).
    \item \textbf{Bonus:} using a bonus token in addition to the placeholder token, with the orthogonal loss between the original placeholder token and the bonus
    \item \textbf{Triple Spare: } using three bonus tokens in addition to the placeholder token, with the orthogonal loss between all combinations of the placeholder and bonus tokens (i.e. for one placeholder token and three spares, we would have 12 different orthogonal loss terms)
\end{enumerate}

\subsection{Prompts}
We used a set of prompts for training and a set of prompts for testing. For subjects, the training prompts were similar to the training prompts used in past works, such as \textit{"a photo of a nice \(\{\}\)"} The test set, \textbf{short prompts}, was similar to the prompts used in past works \citep{gal2022image}, such as \textit{"a picture of \(\{\}\) as a policeman"}. For style prompts, we based our training prompts off of the style prompts used in the repo for \citet{dong2023dreamartist}, such as \textit{"a cropped painting, art by \(\{\}\)"}, and based our test prompts off of those used for subjects, such as \textit{"a police officer, art by \(\{\}\)"} Refer to Appendix \ref{appendix-prompts} for a list of all prompts. 

\subsection{Quantitative Results}
Tables \ref{tab:subject-scores} and \ref{tab:style-scoresl} show the quantitative results. For each metric, for each model, we bolded the highest score.  All relevant training hyperparameters are listed in Appendix \ref{appendix-hyper}. Each method is named for the encoder-denoiser combination used and the token strategy used.
\begin{table}[h]
    \centering
    \begin{tabular}{|c| c|c|c|c|c|c|}
    \hline
    Method & CLIP  Sim & CLIP Cons & Cont Sim & Cont Cons & Img Rew & Pro Sim \\
\hline
T5 Trans Default & 0.6048 & 0.5968 &  0.2952 & 0.3085 & 0.2234 & 0.2785 \\
T5 Trans Bonus & \textbf{0.6393} & 0.6427 &  \textbf{0.3054} & \textbf{0.3235} & 0.1217 & 0.276 \\
T5 Trans Triple Bonus & 0.6287 & \textbf{0.644} & 0.2998 & 0.3197 & -0.0625 & 0.2668 \\
T5 Trans Multi 10 & 0.5718 & 0.5581 & 0.2895 & 0.3043 & 0.2604 & \textbf{0.2802} \\
T5 Trans  Multi & 0.5161 & 0.5293 & 0.2804 & 0.3078 & \textbf{0.3046} & 0.2772 \\
t5 Trans Negative & 0.4884 & 0.527 & 0.2745 & 0.3111 & 0.2758 & 0.2741 \\
\hline
T5 UNet Default & 0.6044 & 0.5859  & 0.2853 & 0.2711 & -0.0436 & 0.2738 \\

T5 UNet Bonus & 0.622 & 0.6201  & 0.2885 & 0.2734 & -0.0686 & 0.269 \\
T5 UNet Triple Bonus & \textbf{0.6575} & \textbf{0.6493} & \textbf{0.2991} & \textbf{0.2904} & -0.2908 & 0.2635 \\
T5 UNet Multi 10  & 0.4747 & 0.5157 & 0.2551 & 0.2482 & 0.2796 & 0.2728 \\
T5 UNet  Multi & 0.4679 & 0.5187 & 0.249 & 0.2414 & \textbf{0.2554} & 0.2707 \\
T5 UNet  Negative & 0.5081 & 0.5435 & 0.2621 & 0.2586 & 0.1888 & \textbf{0.2789} \\
\hline
CLIP UNet Default & 0.7801 & 0.7999 & 0.3149 & 0.2963 & -1.9338 & 0.2032 \\
CLIP UNet Bonus & 0.7696 & 0.7819 & 0.3201 & 0.3015 & -1.624 & 0.2169 \\
CLIP UNet Triple Bonus & 0.8045 & 0.8224 & 0.3471 & 0.335 & -1.8263 & 0.2101 \\
CLIP UNet Multi 10 & \textbf{0.8307} & \textbf{0.8605} & \textbf{0.4019} & \textbf{0.4228} & -1.9432 & 0.2057 \\
CLIP Unet Multi & 0.7959 & 0.7956 & 0.3672 & 0.3542 & -1.2943 & 0.2301 \\
CLIP Unet  Negative & 0.5382 & 0.5805 & 0.2589 & 0.2461 & \textbf{0.1165} & \textbf{0.2756} \\

    \hline
    \end{tabular}
    \caption{Subject Scores}
    \label{tab:subject-scores}
\end{table}


\begin{table}[h]
    \centering
    \begin{tabular}{|c| c|c|c|c|c|c|}
    \hline
    Method & CLIP  Sim & CLIP Cons & Sty Sim & Sty Cons & Img Rew & Pro Sim   \\
    \hline
    T5 Trans Default  & 0.5837 & 0.6053 & 0.2338 & 0.2536 & 0.493 & 0.2867 \\
T5 Trans Bonus & \textbf{0.6325} & \textbf{0.6431} & \textbf{0.3152} & \textbf{0.3431} & 0.3458 & 0.2802 \\
T5 Trans  Triple Bonus & 0.6167 & 0.6202 & 0.264 & 0.2957 & 0.6289 & \textbf{0.2955} \\
T5 Trans  Multi 10 & 0.5821 & 0.6226 & 0.2295 & 0.2475 & 0.6131 & 0.2882 \\
T5 Trans  Multi 50 & 0.5571 & 0.6068 & 0.19 & 0.2148 & \textbf{0.7548} & 0.293 \\
T5 Trans  Negative & 0.5411 & 0.599 & 0.166 & 0.1973 & 0.6017 & 0.2912 \\
\hline
T5 UNet Default  & 0.5945 & 0.6228 & 0.2825 & 0.3815 & 0.8184 & 0.2795 \\
T5 UNet Bonus & \textbf{0.6154} & \textbf{0.6332} & \textbf{0.2988} & \textbf{0.4135} & 0.4593 & 0.275 \\
T5 UNet Triple Bonus & 0.61 & 0.6326 & 0.2889 & 0.4007 & 0.7614 & 0.2837 \\
T5 UNet Multi 10 & 0.5523 & 0.6113 & 0.2108 & 0.2936 & \textbf{1.1392} & 0.292 \\
T5 UNet Multi 50 & 0.5546 & 0.616 & 0.2017 & 0.2865 & 1.0833 & \textbf{0.2915} \\
T5 UNet Negative & 0.5507 & 0.6049 & 0.2066 & 0.2731 & 1.0446 & 0.2903 \\
\hline
CLIP UNet Default  & 0.8003 & 0.8302 & 0.5693 & 0.7011 & -1.4752 & 0.2151 \\
CLIP UNet  Bonus & 0.8284 & 0.8496 & 0.6262 & 0.732 & -1.4739 & 0.2204 \\
CLIP UNet Triple Bonus & 0.8361 & 0.865 & 0.6707 & 0.7651 & -1.5033 & 0.2168 \\
CLIP UNet Multi 10 & \textbf{0.8835} & \textbf{0.9001} & \textbf{0.7632} & \textbf{0.825} & -1.7364 & 0.2137 \\
CLIP UNet Multi 50 & 0.8776 & 0.8821 & 0.7333 & 0.7627 & -1.6204 & 0.2186 \\
CLIP UNet Negative & 0.6177 & 0.5998 & 0.292 & 0.2952 & \textbf{0.6837} & \textbf{0.2986} \\
\hline
    \end{tabular}
    \caption{Style Scores}
    \label{tab:style-scoresl}
\end{table}

Across all models, the use of the bonus token(s) improves content/style similarity and consistency scores compared to the default, at the expense of lower prompt similarity. This reflects the comparison of methods as shown in \citep{avrahami2024chosenoneconsistentcharacters}, where different subject-to-image methods often exist on a Pareto frontier where higher prompt similarity comes at the expense of lower consistency and vice versa. 



\subsection{Visual Results} 
We display some visual results. Each image was generated with the same prompt. Subjectively, we find that the CLIP UNet tends to "overfit" and sometimes ignore the prompt completely, which is consistent with quantitative results where using the T5 text encoder improved Image Reward and Prompt Similarity..
\clearpage
\begin{figure}[h]
    \centering
    \includegraphics[scale=0.45]{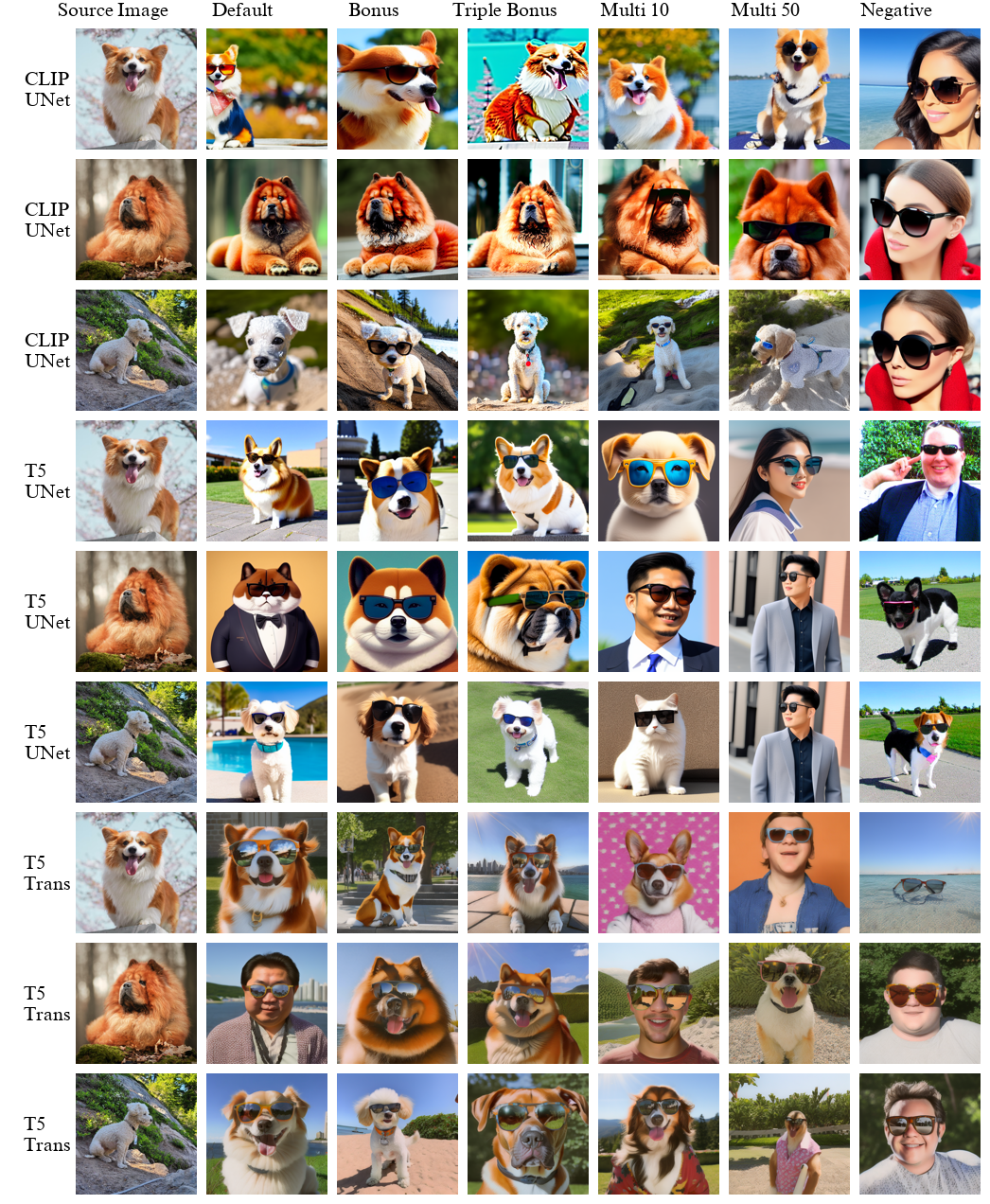}
    \caption{Subject Images, generated with caption "a photo of  \{\} wearing sunglasses"}
    \label{fig:personal-grid}
\end{figure}
\clearpage
\begin{figure}[h]
    \centering
    \includegraphics[scale=0.45]{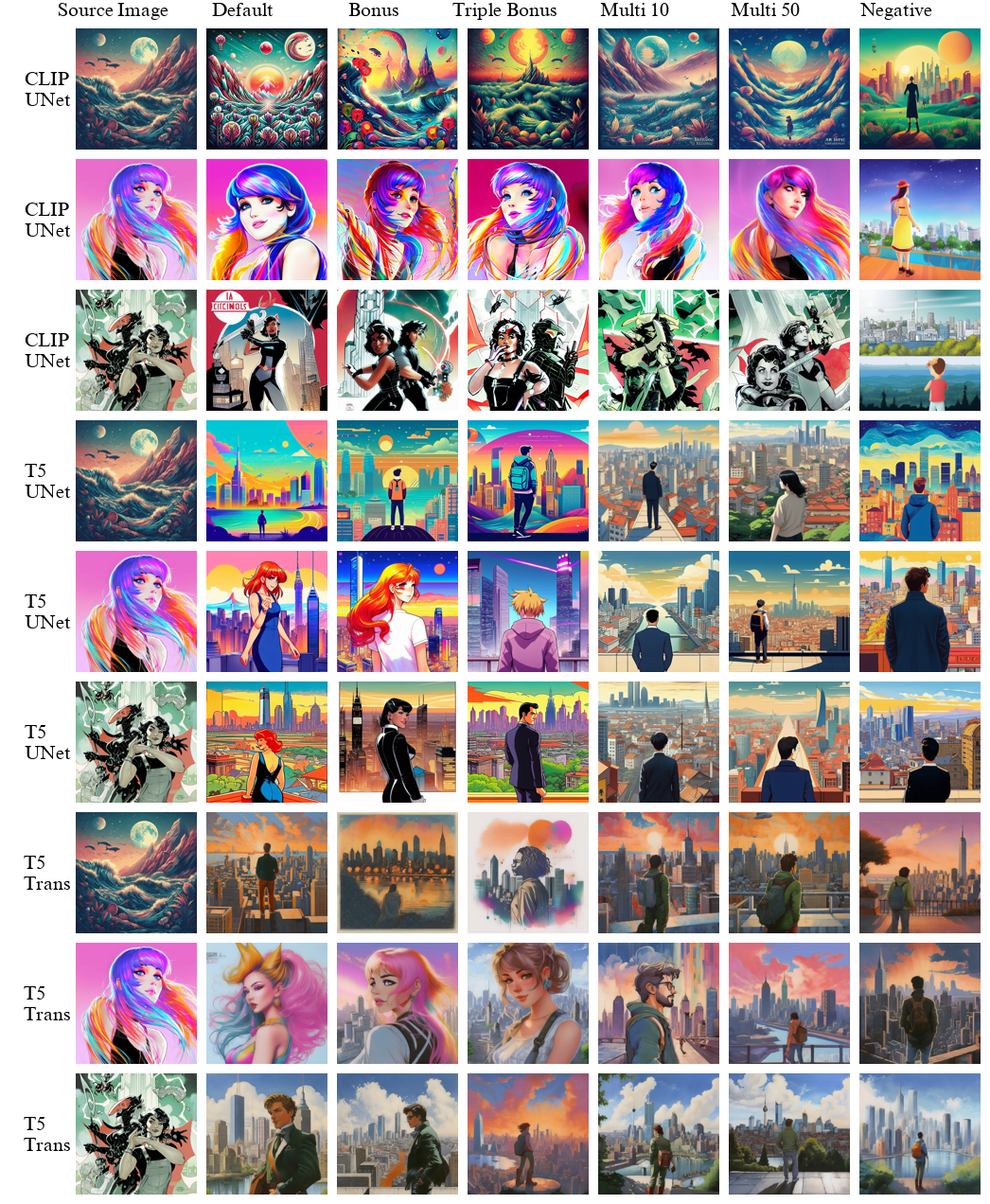}
    \caption{Style Images, generated with caption "a person with a city in the background, art by \{\}" }
    \label{fig:style-grid}
\end{figure}

\section{Conclusion}
To summarize, we observed that textual inversion for personalizing diffusion models was wedded to the UNet architecture, so we experimented with textual inversion that relied on the vision transformer instead, and used BRAT to optimize the tokens for such. We saw each of our contributions reflected a movement along the prompt similarity-content adherence pareto frontier. Further work could entail expanding this approach to more alternatives to the UNet, or longer training times on larger text encoders that may capture richer, more meaningful information in their embedding space.

\subsubsection*{Broader Impact Statement}
Many people are worried about the effects of generative AI. By creating art, this technology encroaches on an area once solely occupied by humans. Companies have faced criticism for potentially using AI, and many creatives, such as screenwriters and actors, have expressed concerns about the security of their jobs. However, AI can assist humans by enhancing efficiency, providing inspiration, and generating new ideas. The future of copyright protection for AI-generated art remains uncertain, as current laws are based on the principle that creative works originate from human authors. To mitigate harm and maximize benefits for everyone, clear and consistent policies from governments, industries, and academic groups will be necessary.

\section{Assistance}

\subsection*{Author Contributions}
This work was done without any outside assistance or collaboration.

\subsection*{Acknowledgements}
Rutgers Office of Advanced Research Computing kindly provided the computing infrastructure to run the experiments. A special thanks goes out to Dr. Ahmed Elgammal and Dr. Eugene White for their past advice prior to this work.

\bibliography{main}
\bibliographystyle{tmlr}

\appendix

\section{Appendix}
\subsection{Style Dataset}\label{appendix-style}
\begin{figure}
    \centering
    \includegraphics[scale=0.6]{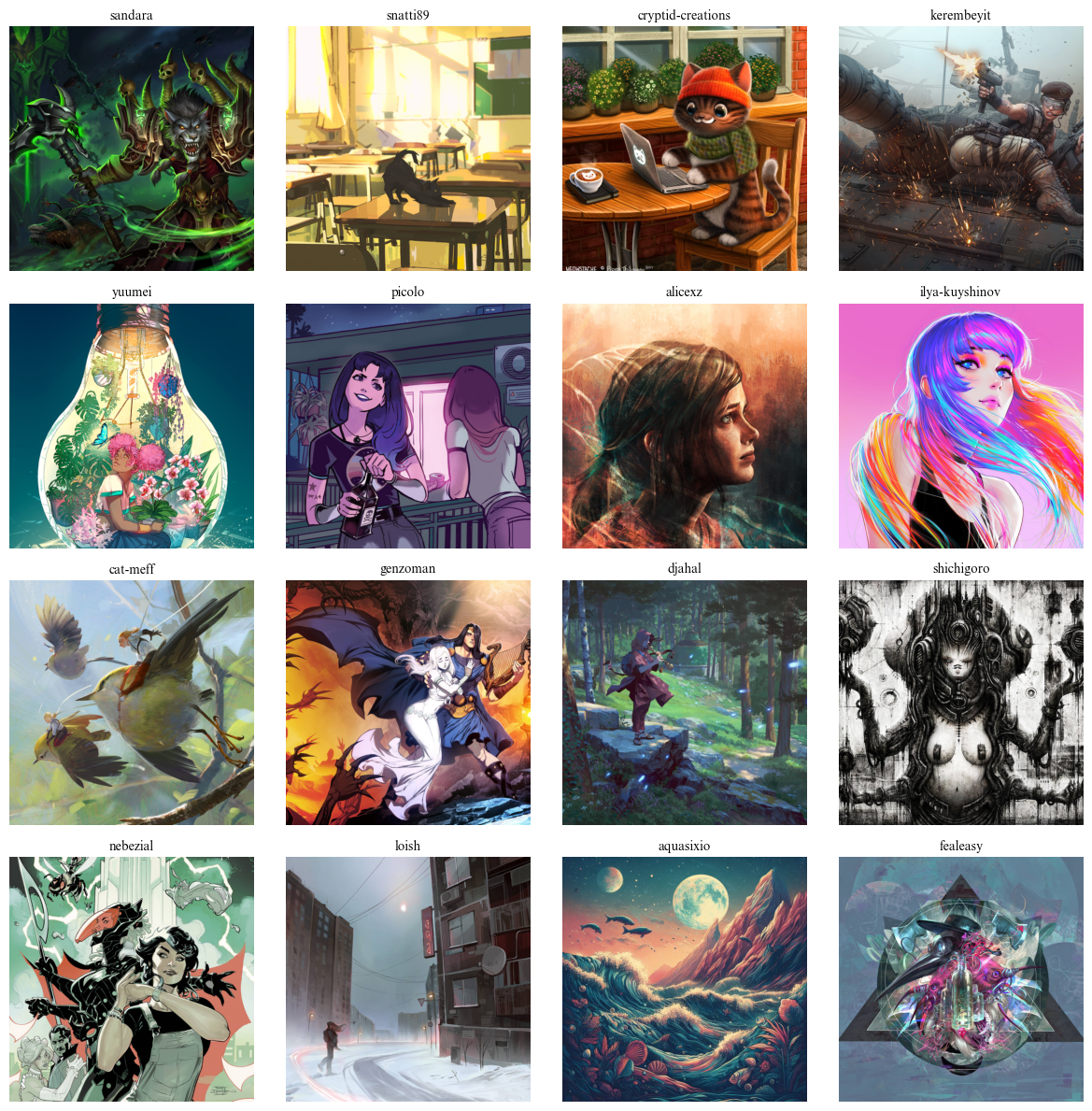}
    \caption{Source Images, labeled by their deviantart usernames}
    \label{fig:src}
\end{figure}

The style data can be found at \url{https://huggingface.co/datasets/jlbaker361/stylization}. A list of the artists, and links to their profiles across whatever platforms could be found, is as follows:
\begin{enumerate}
    \item \textbf{Lois van Baarle} Deviantart: \url{https://www.deviantart.com/loish}; Instagram: \url{https://www.instagram.com/loisvb}; Personal Website: \url{https://loish.net/}
    \item \textbf{Kerem Beyit} Deviantart: \url{https://www.deviantart.com/kerembeyit}; Instagram: \url{https://www.instagram.com/kerembeyit}
    \item \textbf{sandara} Deviantart: \url{https://www.deviantart.com/sandara}
    \item \textbf{yuumei} Deviantart: \url{https://www.deviantart.com/yuumei}; Personal Website: \url{https://www.yuumeiart.com/}
    \item \textbf{Gabriel Picolo} Deviantart: \url{https://www.deviantart.com/picolo-kun}; Instagram: \url{https://www.instagram.com/_picolo/}
    \item \textbf{Ilya Kuvshinov} Deviantart: \url{https://www.deviantart.com/kuvshinov-ilya}; Instagram: \url{https://www.instagram.com/kuvshinov_ilya/}
    \item \textbf{Cryptid Creations} Deviantart: \url{https://www.deviantart.com/cryptid-creations}
    \item \textbf{alicexz} Deviantart: \url{https://www.deviantart.com/alicexz}
    \item \textbf{Atey Ghailan} Deviantart: \url{https://www.deviantart.com/snatti89}; Tumblr: \url{https://snatti.tumblr.com/}; Instagram: \url{https://www.instagram.com/snatti89/}
    \item \textbf{cat-meff} Deviantart: \url{https://www.deviantart.com/cat-meff}
    \item \textbf{Gonzalo Ordonez Arias} Deviantart: \url{https://www.deviantart.com/genzoman}; Instagram: \url{https://www.instagram.com/mrgenzoman/}; Tumblr: \url{https://www.tumblr.com/genzoman}
    \item \textbf{Geoffroy Thoorens} Deviantart: \url{https://www.deviantart.com/djahal}; Instagram: \url{https://www.instagram.com/djahal/?hl=en}; Personal Website: \url{https://djahalland.com/}
    \item \textbf{Shingo Matsunuma} Deviantart: \url{https://www.deviantart.com/shichigoro756};  Personal Website: \url{https://shichigoro.com/en/home/}
    \item \textbf{Stjepan Sejic} Deviantart: \url{https://www.deviantart.com/nebezial}; 
    \item \textbf{Cyril Rolando} Deviantart: \url{https://www.deviantart.com/aquasixio}; Instagram: \url{https://www.instagram.com/aquasixio/?hl=en}; Tumblr: \url{https://cyrilrolando.tumblr.com/}
    \item \textbf{Sophia von Yhlen} Deviantart: \url{https://www.deviantart.com/fealasy}; Instagram: \url{https://www.instagram.com/fealasy/}
\end{enumerate}
These images are shown in figure \ref{fig:src}

\subsection{Prompts}\label{appendix-prompts}
Tables \ref{tab:subject-prompts} and \ref{tab:style-prompts} list the training prompts used for the subjects and styles, respectively. Tables \ref{tab:subject_evaluations} and \ref{tab:style_evaluationl} list the evaluation prompts for subjects and styles, respectively.

\begin{table}[h]
    \centering
    \begin{tabular}{|c|c|c|}
    \hline
        a photo of a \(\{\}\) & a rendering of a \(\{\}\) & a cropped photo of the \(\{\}\) \\
        \hline
the photo of a \(\{\}\) & a photo of a clean \(\{\}\) & a photo of a dirty \(\{\}\) \\
\hline
a dark photo of the \(\{\}\) & a photo of my \(\{\}\) & a photo of the cool \(\{\}\) \\
\hline
a close-up photo of a \(\{\}\) & a bright photo of the \(\{\}\) & a cropped photo of a \(\{\}\) \\
\hline
a photo of the \(\{\}\) & a good photo of the \(\{\}\) & a photo of one \(\{\}\) \\
\hline
a close-up photo of the \(\{\}\) & a rendition of the \(\{\}\) & a photo of the clean \(\{\}\) \\
\hline
a rendition of a \(\{\}\) & a photo of a nice \(\{\}\) & a good photo of a \(\{\}\) \\
\hline
a photo of the nice \(\{\}\) & a photo of the small \(\{\}\) & a photo of the weird \(\{\}\) \\
\hline
a photo of the large \(\{\}\) & a photo of a cool \(\{\}\) & a photo of a small \(\{\}\) \\
\hline
    \end{tabular}
    \caption{Subject Prompts}
    \label{tab:subject-prompts}
\end{table}

\begin{table}[h]
    \centering
    \begin{tabular}{|c|c|c|}
    \hline
a painting, art by \(\{\}\) & a rendering, art by \(\{\}\) & a cropped painting, art by \(\{\}\) \\
\hline
the painting, art by \(\{\}\) & a clean image, art by \(\{\}\) & a dirty image, art by \(\{\}\) \\
\hline
a dark image, art by \(\{\}\) & an image, art by \(\{\}\) & a cool picture, art by \(\{\}\) \\
\hline
a close-up picture, art by \(\{\}\) & a bright picture, art by \(\{\}\) & a cropped picture, art by \(\{\}\) \\
\hline
a good painting, art by \(\{\}\) & a close-up painting, art by \(\{\}\) & a rendition, art by \(\{\}\) \\
\hline
a nice painting, in the style of \(\{\}\) & a small painting, in the style of \(\{\}\) & a weird painting, in the style of \(\{\}\) \\
\hline
a large painting, in the style of \(\{\}\) & & \\
\hline
    \end{tabular}
    \caption{Style Prompts}
    \label{tab:style-prompts}
\end{table}

\begin{table}[h]
    \centering
    \begin{tabular}{|c|c|c|}
    \hline
a photo of  \(\{\}\) at the beach & a photo of  \(\{\}\) in the jungle \\ \hline
a photo of  \(\{\}\) in the snow & a photo of  \(\{\}\) in the street \\ \hline
a photo of  \(\{\}\) with a city in the background & a photo of  \(\{\}\) with a mountain in the background \\ \hline
a photo of  \(\{\}\) with the Eiffel Tower in the background & a photo of  \(\{\}\) near the Statue of Liberty \\ \hline
a photo of  \(\{\}\) near the Sydney Opera House & a photo of  \(\{\}\) floating on top of water \\ \hline
a photo of  \(\{\}\) eating a burger & a photo of  \(\{\}\) drinking a beer \\ \hline
a photo of  \(\{\}\) wearing a blue hat & a photo of  \(\{\}\) wearing sunglasses \\ \hline
a photo of  \(\{\}\) playing with a ball & a photo of  \(\{\}\) as a police officer \\ \hline
    \end{tabular}
    \caption{Subject Evaluation Prompt}
    \label{tab:subject_evaluations}
\end{table}

\begin{table}[h]
    \centering
    \begin{tabular}{|c|c|}
         \hline
the beach, art by \(\{\}\) & the jungle, art by \(\{\}\) \\ \hline
the snow, art by \(\{\}\) & the street, art by \(\{\}\) \\ \hline
a person with a city in the background, art by \(\{\}\) & a person with a mountain in the background, art by \(\{\}\) \\ \hline
the Eiffel Tower, art by \(\{\}\) & the Statue of Liberty, art by \(\{\}\) \\ \hline
the Sydney Opera House, art by \(\{\}\) & person floating on top of water, art by \(\{\}\) \\ \hline
eating a burger, art by \(\{\}\) & drinking a beer, art by \(\{\}\) \\ \hline
wearing a blue hat, art by \(\{\}\) & wearing sunglasses, art by \(\{\}\) \\ \hline
playing with a ball, art by \(\{\}\) & a police officer, art by \(\{\}\) \\ \hline
    \end{tabular}
    \caption{Style Evaluation Prompts}
    \label{tab:style_evaluationl}
\end{table}

\clearpage
\subsection{Additional Images}\label{appendix-images}
\begin{figure}[h]
    \centering
    \includegraphics[scale=0.45]{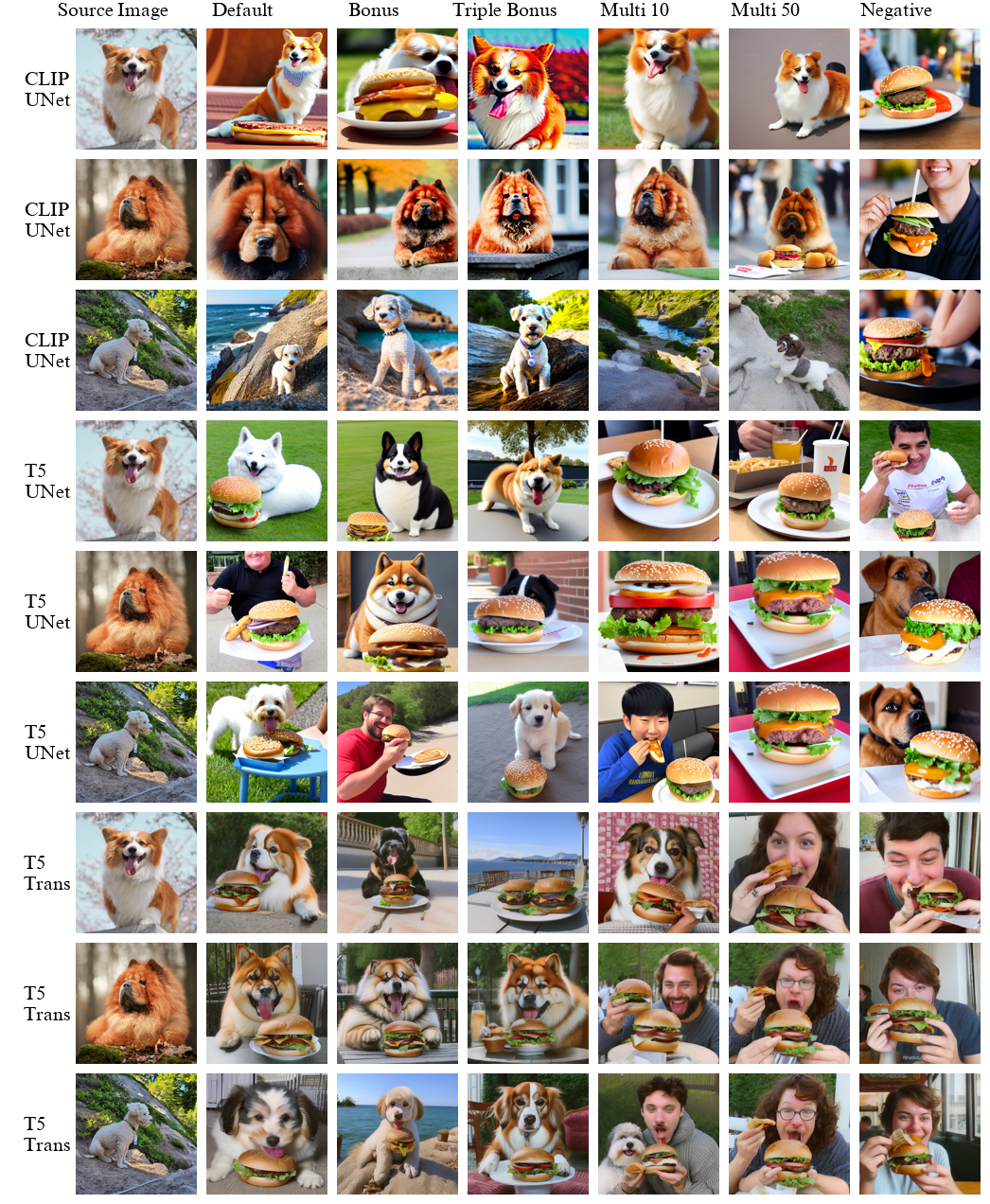}
    \caption{Images generated with the prompt "a photo of  \(\{\}\) eating a burger"}
    \label{fig:burger}
\end{figure}
\clearpage
\begin{figure}[h]
    \centering
    \includegraphics[scale=.45]{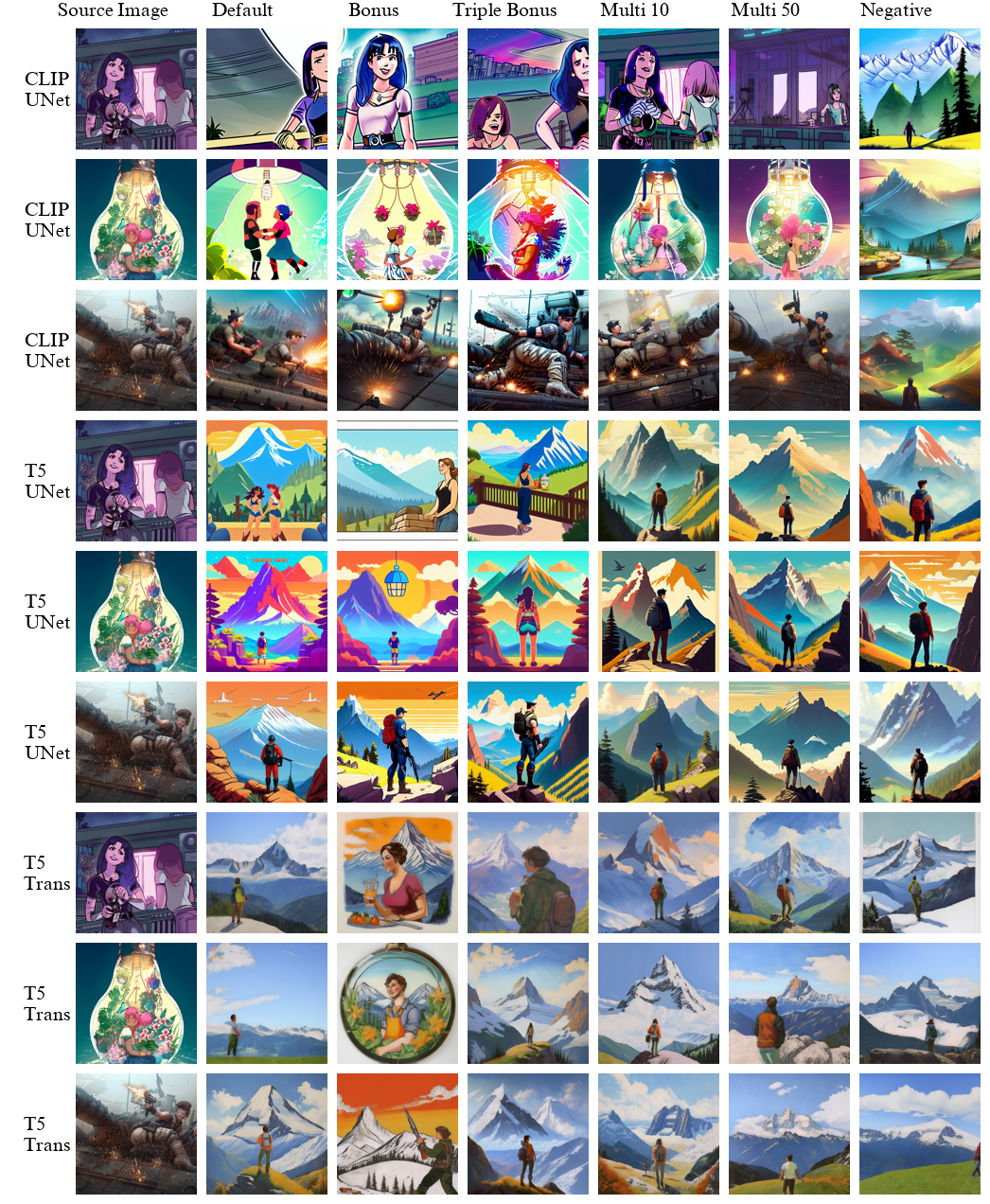}
    \caption{Images generated with prompt "a person with a mountain in the background, art by \{\}"}
    \label{fig:mountain}
\end{figure}
\subsection{Training}
All experiments were done using an A100 GPU with 40GB RAM. All code was written in Python 3.11, leveraging libraries such as PyTorch \citep{paszke2019pytorchimperativestylehighperformance}, Diffusers \citep{von-platen-etal-2022-diffusers}, TRL \citep{vonwerra2022trl}, Accelerate \citep{accelerate2022} and Wandb \citep{wandb2020}. 
\subsubsection{Hyperparameters}\label{appendix-hyper}
Training Hyperparameters are listed in table \ref{tab:hp-table}. Training for the style and subject datasets used all of the same hyperparameters except for the number of epochs.
\begin{table}[h]
    \centering
    \begin{tabular}{|c|c|}
    \hline
    Parameter & Value \\
    \hline
       Epochs (Subjects)  & 250 \\
       Epochs (Styles) & 500 \\
       Learning Rate & 0.08 \\
       Gradient Accumulation Steps & 8 \\
       Batch Size & 1 \\
       Spare \(\lambda\) & 0.01 \\
       Noise Scheduler & DDPM \\
       Max Gradient Norm & 10.0 \\
       \hline
    \end{tabular}
    \caption{Hyperparameters}
    \label{tab:hp-table}
\end{table}
\subsection{Failed Methods}\label{appendix-fail}
\begin{figure}[h]
    \centering
    \includegraphics[]{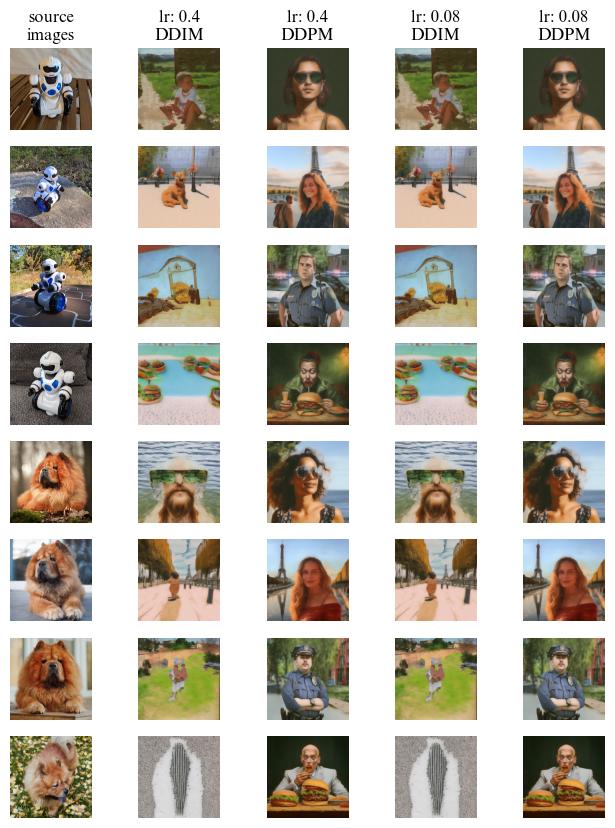}
    \caption{PixArt }
    \label{fig:ugly-pixart}
\end{figure}

\begin{figure}[h]
    \centering
    \includegraphics[]{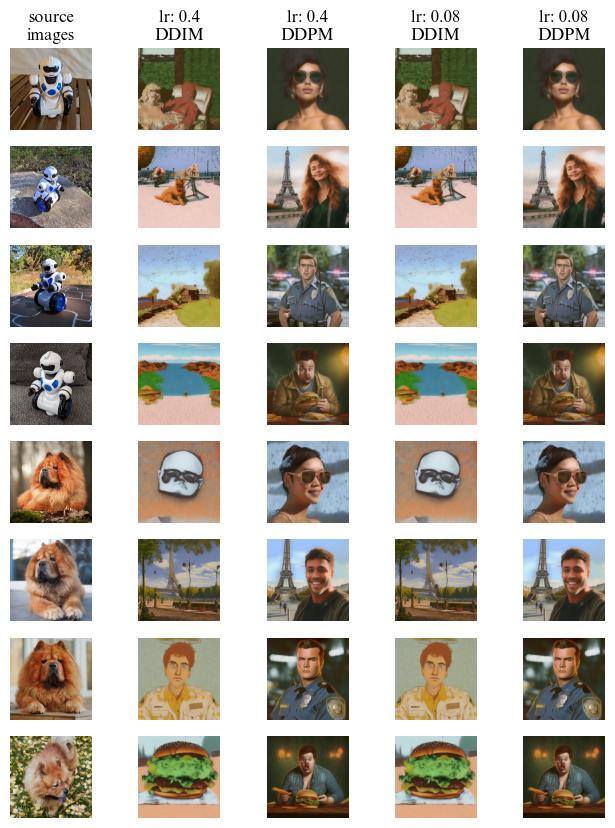}
    \caption{PixArt (With Bonus Token) }
    \label{fig:ugly-pixart-spare}
\end{figure}

\begin{figure}[h]
    \centering
    \includegraphics[]{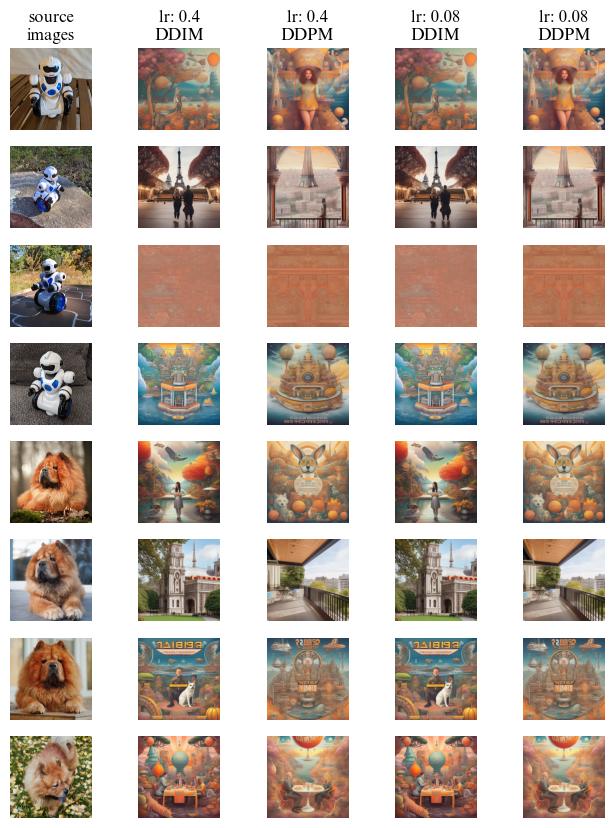}
    \caption{Llama}
    \label{fig:ugly-llama}
\end{figure}

\begin{figure}[h]
    \centering
    \includegraphics[]{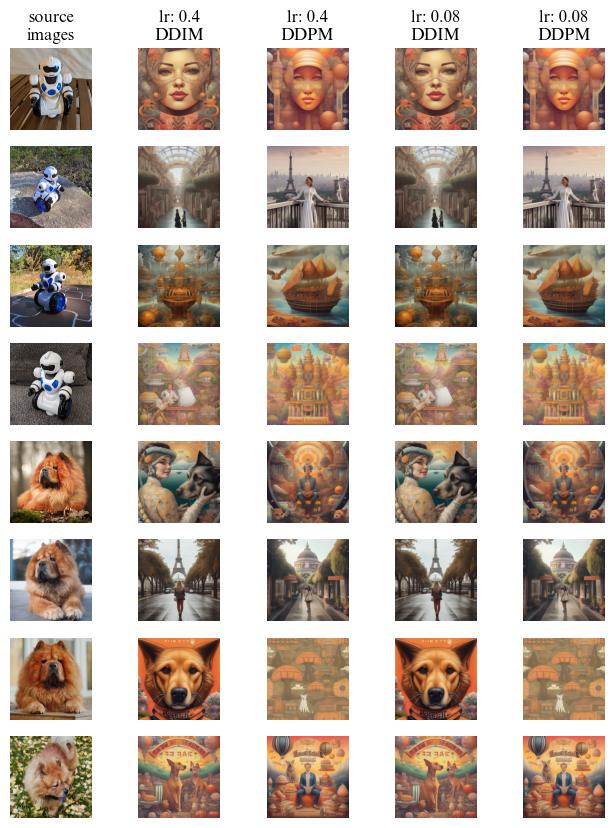}
    \caption{Llama (With Bonus Token)}
    \label{fig:ugly-llama-spare}
\end{figure}

We briefly experimented with large text encoders with more than a billion parameters. The \textbf{PixArt-XL-2-512x512} text encoder had roughly 4.7 billion parameters, and the \textbf{Llama-2-7b-hf} had roughly 6.7 billion parameters. We found these very difficult to train, failing to capture the subjects at all. We attempted training with both traditional textual inversion and using the spare token. We used 750 epochs, instead of 250. Nonetheless, we found unsatisfying results. Figures \ref{fig:ugly-pixart}, \ref{fig:ugly-pixart-spare}, \ref{fig:ugly-llama} and \ref{fig:ugly-llama-spare} show some examples. The left-most column of each figure shows the source images from the personalization dataset, and each row shows the image generated from the same prompt, where each column denotes a different combination of learning rate for training (0.08 or 0.4) and scheduler (DDIM \citep{song2022denoisingdiffusionimplicitmodels} or DDPM \citep{ho2020denoisingdiffusionprobabilisticmodels}).

\end{document}